\pdfoutput=1

\documentclass[11pt]{article}

\usepackage[final]{acl}

\usepackage{times}
\usepackage{latexsym}

\usepackage[T1]{fontenc}

\usepackage[utf8]{inputenc}

\usepackage{microtype}

\usepackage{inconsolata}

\usepackage{graphicx}

\usepackage{tabularx} 
\usepackage{array}
\usepackage{multirow}
\usepackage{booktabs}
\usepackage{graphicx} 
\usepackage{url} 
\usepackage{amsthm,amsmath,amssymb} 
\usepackage{mathrsfs} 
\usepackage{bm} 
\usepackage{pifont} 
\usepackage{float}
\usepackage{lipsum}
\usepackage{soul}
\usepackage{caption}
\usepackage{subcaption}
\usepackage{enumitem}
\usepackage{bbm}
\usepackage{dsfont}
\usepackage{color}
\usepackage{colortbl}
\usepackage{xcolor}
\definecolor{deepred}{RGB}{180,0,0} 
\definecolor{customred}{RGB}{170, 20, 20} 
\definecolor{customgreen}{RGB}{20, 170, 20} 
\usepackage{xspace}

\title{Instruction-Tuning Data Synthesis from Scratch via Web Reconstruction}

\author{
Yuxin Jiang$^{1,2}$\thanks{~~Work done during the internship at Huawei Noah's Ark Lab. Data and code: \url{https://github.com/YJiangcm/WebR}.}, 
Yufei Wang$^{3}$, 
Chuhan Wu$^{3}$, 
Xinyi Dai$^{3}$, 
\textbf{Yan Xu}$^{3}$\textbf{,} 
\textbf{Weinan Gan}$^{3}$\textbf{,} 
\\
\textbf{Yasheng Wang}$^{3}$\textbf{,} 
\textbf{Xin Jiang}$^{3}$\textbf{,} 
\textbf{Lifeng Shang}$^{3}$\textbf{,} 
\textbf{Ruiming Tang}$^{3}$\textbf{,} 
\textbf{Wei Wang}$^{1,2}$
\\
The Hong Kong University of Science and Technology (Guangzhou)$^1$, \\
The Hong Kong University of Science and Technology$^2$, Huawei Noah’s Ark Lab$^3$ \\
yjiangcm@connect.ust.hk, wang.yufei1@huawei.com, weiwcs@ust.hk 
}

\begin{document}
\maketitle
\begin{abstract}
The improvement of LLMs’ instruction-following capabilities depends critically on the availability of high-quality instruction-response pairs.
While existing automatic data synthetic methods alleviate the burden of manual curation, they often rely heavily on either the quality of seed data or strong assumptions about the structure and content of web documents.
To tackle these challenges, we propose \textbf{Web Reconstruction} (WebR), a fully automated framework for synthesizing high-quality instruction-tuning (IT) data directly from raw web documents with minimal assumptions.
Leveraging the inherent diversity of raw web content, we conceptualize \textit{web reconstruction} as an instruction-tuning data synthesis task via a novel dual-perspective paradigm—\textit{Web as Instruction} and \textit{Web as Response}—where each web document is designated as either an instruction or a response to trigger the reconstruction process.
Comprehensive experiments show that datasets generated by WebR outperform state-of-the-art baselines by up to 16.65\% across four instruction-following benchmarks.
Notably, WebR demonstrates superior compatibility, data efficiency, and scalability, enabling enhanced domain adaptation with minimal effort.
\end{abstract}

\section{Introduction}
Large language models (LLMs)~\cite{brown2020fewshot, 2023gpt4, llama3modelcard} have become integral across a myriad of applications, demonstrating exceptional performance on diverse tasks by effectively following instructions~\cite{openai2022chatgpt, 2023gpt4}.
Their remarkable performance largely stems from supervised fine-tuning (SFT)~\cite{wei2022finetuned, mishra-etal-2022-cross} on instruction-response pairs.
This process empowers LLMs to produce customized outputs when provided with specific instructions, facilitating their adaptation to novel tasks without prior exposure.

\begin{figure}[!t]
\centering
\includegraphics[width=\linewidth]{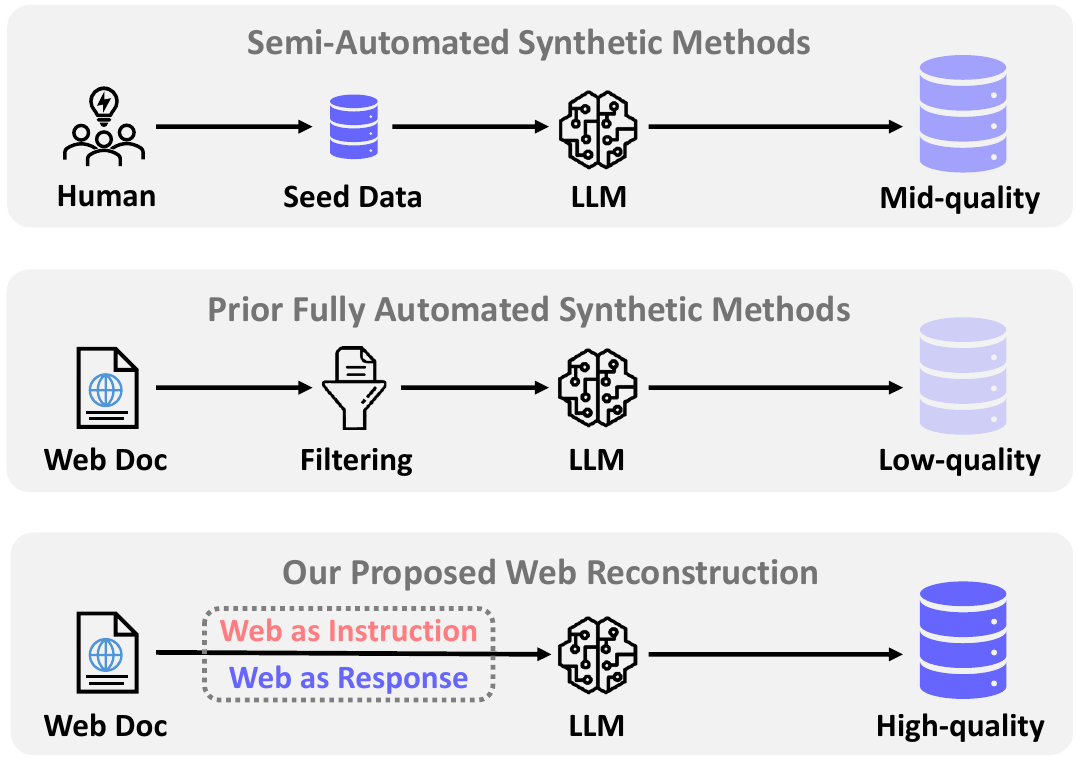}
\caption{
Our proposed Web Reconstruction method surpasses previous techniques by being (1) fully automated, eliminating the need for manual intervention or seed data; (2) minimally reliant on assumptions about the structure and content of web documents; and (3) capable of generating high-quality IT data.
}
\label{fig:intro}
\end{figure}

A fundamental challenge in advancing the instruction-following capabilities of LLMs lies in the collection of high-quality instruction-tuning (IT) data.
Early approaches primarily rely on human experts to manually generate and curate IT data~\cite{wang-etal-2022-super, DatabricksBlog2023DollyV2}, which is both time-intensive and resource-heavy.
To mitigate these limitations, \textbf{Semi-Automated Synthetic Methods}~\cite{wang-etal-2023-self-instruct, alpaca, xu2024wizardlm} leverage LLMs to expand small, human-annotated seed datasets using few-shot prompting techniques.
While effective, the performance of these methods is highly sensitive to prompt engineering and the careful selection of seed examples~\cite{xu2024magpie}.
More recently, \textbf{Fully Automated Synthetic Methods}, such as WebInstruct~\cite{yue2024mammoth2} and instruction backtranslation~\cite{li2024selfalignment}, have emerged as scalable alternatives that eliminate human involvement by synthesizing IT data based on web-scraped documents.
These methods, however, often operate under strong assumptions about the structure and content of raw web data, such as the availability of explicit question-answer pairs or minimal irrelevant content.
Consequently, they can only handle a limited scope of web documents, restricting their diversity and leading to suboptimal performance across various tasks.



To overcome these limitations, we propose \textbf{Web Reconstruction (WebR)}—a novel framework that synthesizes high-quality IT data from raw web documents \textbf{with minimal assumptions on web} and \textbf{no reliance on human annotations}, enabling broader adaptability and improved performance.
Unlike backtranslation~\cite{li2024selfalignment}, which directly treats web content as a response, or WebInstruct~\cite{yue2024mammoth2}, which extracts QA pairs, WebR introduces a novel paradigm by \textbf{conceptualizing web reconstruction as an instruction-tuning data synthesis task}.
At its core, WebR aims to transform raw, noisy web documents into human-preferred, response-like outputs through a dual-perspective paradigm. Each web document is designated as either an instruction or a response, triggering the reconstruction process:
(1) \textit{Web as Instruction} introduces a first-of-its-kind \textbf{web rewriting} approach in IT data synthesis, where raw web document is concatenated with a synthesized rewrite request to serve as a complete instruction;
(2) \textit{Web as Response} enhances backtranslation by introducing a novel rollout and refinement process, mitigating reliance on strong web content assumptions.
Crucially, we show that these two perspectives are \textbf{complementary} (refer to Table \ref{tab:ablation}): \textit{Web as Instruction} enhances reasoning and understanding tasks, while \textit{Web as Response} improves instruction-following and question-answering tasks.

We apply WebR to the Llama3-70B-Instruct and GPT-4o-mini models, creating two 100k-sample IT datasets: WebR-Basic and WebR-Pro.
To validate their effectiveness, we train various LLMs, including Llama3-8B-base and Qwen2.5-1.5/3/7/14B-base, and evaluate them on over ten widely used benchmarks.
Our experiments provide key contributions and insights into IT data synthesis:
\begin{itemize}
    \item \textbf{Efficacy:} WebR is the first web-based IT synthetic method to consistently surpass current IT datasets with human annotations;
    \item \textbf{Compatibility:} Merging WebR with existing IT datasets yields further performance gains;
    \item \textbf{Data Efficiency:} The performance of WebR improves linearly relative to the logarithmic growth of training data;
    \item \textbf{Scalability:} WebR scales with LLM size, consistently boosting larger models;
    \item \textbf{Domain Adaptability:} WebR achieves domain adaption by simply adjusting the proportion of source web documents.
\end{itemize}




\begin{figure*}[!t]
\centering
\includegraphics[width=\linewidth]{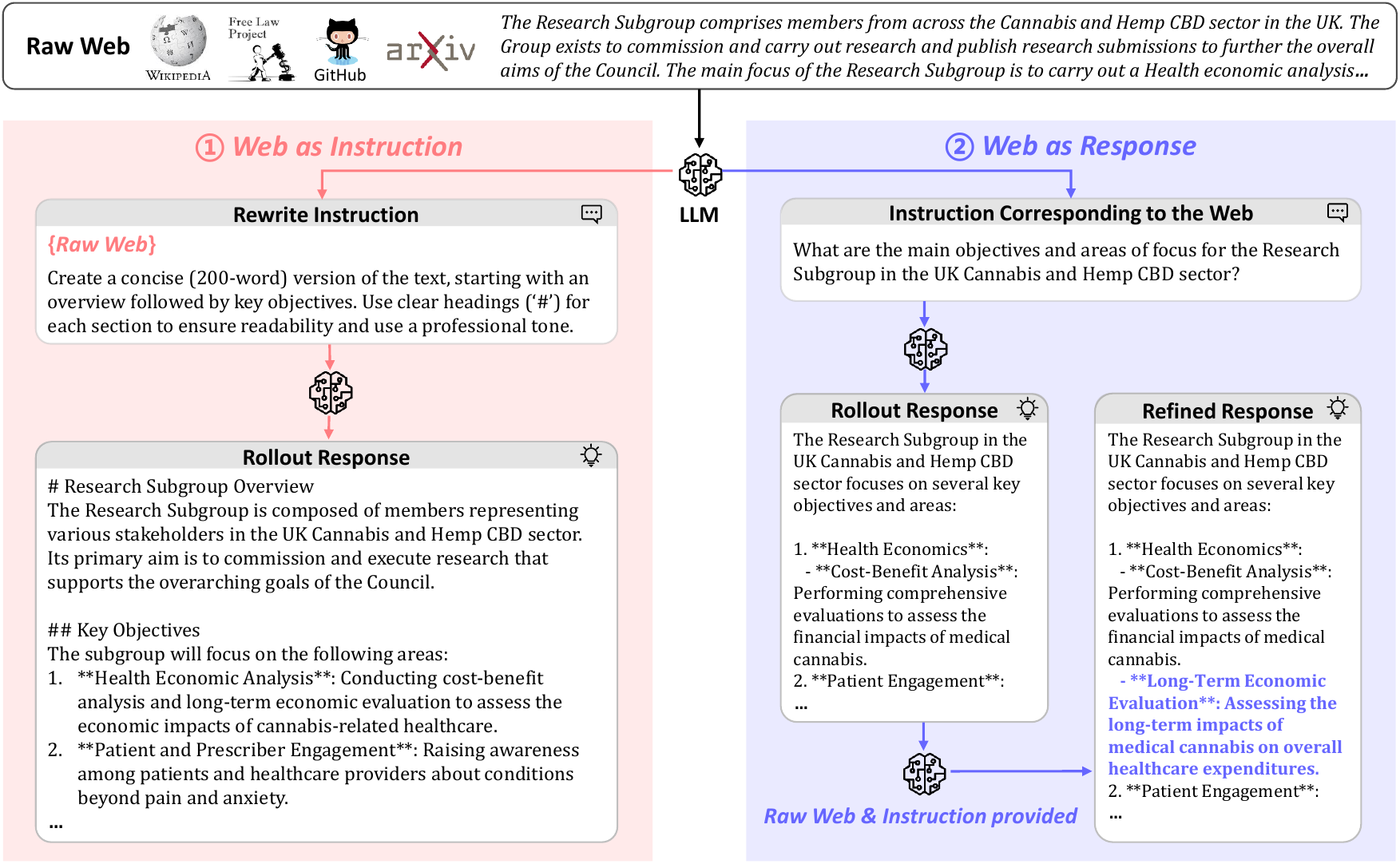}
\caption{
Overview of the proposed \textbf{Web Reconstruction} (WebR) framework.
Leveraging an off-the-shelf LLM, WebR transforms raw web documents into high-quality instruction-response pairs.
It strategically assigns each document as either an instruction or a response to trigger the process of web reconstruction.
}
\label{fig:method}
\end{figure*}

\section{Related Work}


The synthesis of high-quality instruction-tuning (IT) data~\cite{zhou2023lima, jiang2023lion, DBLP:conf/acl/JiangWWZZGLJSTL24, xu2024wizardlm} can be broadly classified into three main categories:

\paragraph{Human-Crafted Method}
primarily involves employing professionals to create instructions, as seen in datasets like SUPER-NI~\cite{wang-etal-2022-super} and DOLLY~\cite{DatabricksBlog2023DollyV2}.
While these datasets offer high-quality content, their size is constrained by the significant costs associated with manual creation.
Alternatively, approaches like ShareGPT~\cite{vicuna2023} and WildChat~\cite{zhao2024wildchat} leverage user interaction logs with LLMs to collect human-generated instructions.
However, this method risks incorporating toxic or undesirable content~\cite{zhao2024wildchat}.

\paragraph{Semi-Automated Synthetic Method}
uses LLMs to generate synthetic IT datasets by starting with a small set of human-annotated seed data and expanding them through few-shot prompting.
Notable methods include Self-Instruct~\cite{wang-etal-2023-self-instruct}, Alpaca~\cite{alpaca}, and Evol-Instruct~\cite{xu2024wizardlm}.
While these techniques enable large-scale data generation, the diversity of the synthesized data is often constrained by the quality and variety of seed examples~\cite{li2024synthetic}.

\paragraph{Fully Automated Synthetic Method}
utilizes LLMs to synthesize IT data from scratch, drawing from web-scraped documents.
For instance, WebInstruct~\cite{yue2024mammoth2} extracts question-answer (QA) pairs from web documents to construct instruction-response datasets.
Nevertheless, this approach depends on the explicit presence of QA pairs within the raw web corpus, which is not always guaranteed.
Similarly, backtranslation~\cite{li2024selfalignment, nguyen2024betteralignmentinstructionbackandforth} treats web documents as natural responses and employs LLMs to infer the corresponding latent user instructions.
However, web documents often contain irrelevant content or unsuitable expressions, making them suboptimal as response candidates.

\section{Web Reconstruction}
Prior fully automated synthetic methods often rely on strong assumptions about the structure and content of raw web documents—such as the presence of explicit question-answer pairs, minimal irrelevant content, or appropriate expressions—necessitating complex preprocessing steps like retrieval and filtering.
In contrast, we introduce the \textbf{Web Reconstruction} (WebR) framework, which leverages a powerful, off-the-shelf LLM to overcome  these limitations by directly reconstructing unstructured and noisy web content into high-quality, response-like outputs.
As shown in Figure \ref{fig:method}, WebR comprises two core strategies: (1) \textit{Web as Instruction}, where raw web content is concatenated with a synthesized rewrite request to serve as a complete instruction, guiding the generation of a reorganized, coherent response;
(2) \textit{Web as Response}, where a latent instruction is inferred by treating raw web content as a response, enabling reconstruction through the LLM's initial rollout and subsequent refinement.
By adopting this dual-branch approach, WebR efficiently generates high-quality instruction-response pairs, ensuring contextually appropriate outputs while eliminating the need for extensive preprocessing.

\subsection{Web as Instruction}
Raw web documents often contain disorganized or irrelevant information that hinders direct usability.
Even when dealing with well-structured content, further refinement is often required to meet human-preferred formats and stylistic conventions. 
A natural approach to reconstructing web content is to rewrite it according to specific requirements, such as style, format, structure, etc.
To ensure diverse and realistic rewriting scenarios, we leverage a powerful LLM to generate a detailed rewrite request tailored to the original document’s content (See prompt in Figure \ref{fig:l2l_all_prompt}).
The request, along with the raw web content, are concatenated to form a comprehensive instruction.
In addition to whole-document transformations, we further enhance task diversity by randomly (50\% probability) generating rewrite requests that target \textit{specific sections} of the web content rather than the entire document, as shown in Figure \ref{fig:l2l_part_prompt}.
This simulates real-world text manipulation scenarios where users may need to extract and modify only certain portions of a text.
The curated instructions are then processed by the LLM to produce reconstructed web content.
Notably, 
the complexity of rewrite requests naturally encompasses various NLP tasks, such as summarization, information extraction, and semantic understanding.
Addressing these tasks requires LLM to demonstrate advanced reasoning and comprehension abilities, thereby enhancing its proficiency in instruction-following, contextual understanding, and reasoning (as verified in Table \ref{tab:ablation}).

\subsection{Web as Response}
Inspired by instruction backtranslation~\cite{li2024selfalignment}, we propose an alternative approach to reconstruct web content by treating the web as a response.
Specifically, we utilize a LLM to predict a latent instruction for which the raw web content would serve as an ideal response, as illustrated in Figure \ref{fig:s2l_all_prompt}.
To further enhance diversity, specific segments of web content are treated as responses (with a 50\% probability), as depicted in Figure \ref{fig:s2l_part_prompt}.
Unlike traditional back-translation methods, which directly treat latent instructions and raw web content as instruction-response pairs, \textbf{our approach introduces a two-stage refinement process}. First, we generate an initial response by rolling out an LLM prediction for the latent instruction. Next, we refine this response using both the raw web content and the latent instruction to produce a more accurate and comprehensive output, as shown in Figure \ref{fig:refine_prompt}.
The initial rollout ensures that the response exhibits human-like fluency and natural language style, while the subsequent refinement step integrates critical information from the raw web, ensuring that the final response is both precise and thorough. This dual-stage process significantly enhances the LLM's performance in knowledge acquisition and question-answering tasks, as demonstrated by the improvements reported in Table \ref{tab:ablation}.
The generated instruction as well as the refined response are finally paired as IT data.

\subsection{Dataset Construction Details}
Following prior work~\cite{li2024selfalignment, yue2024mammoth2}, we construct our dataset by sampling raw web documents from three diverse and representative domains: 70\% from the English subset of Common Crawl~\cite{together2023redpajama} (general domain), 15\% from OpenWebMath~\cite{paster2024openwebmath} (math domain), and 15\% from GitHub~\cite{together2023redpajama} (code domain).
To enable large-scale creation of diverse synthetic data for various scenarios, we adopt a persona-driven instruction synthesis strategy inspired by \citet{DBLP:journals/corr/abs-2406-20094}.
Initially, an LLM generates personas for the raw web documents (see template in Figure \ref{fig:persona_prompt}), which guide the subsequent instruction synthesis for our proposed Web Reconstruction process.
The ratio of \textit{Web as Instruction} to \textit{Web as Response} is set to 2:1, following insights from the ablation study presented in Table \ref{tab:ablation}.
To enhance diversity and eliminate redundancy, we apply MinHash~\cite{broder1997resemblance} deduplication based on n-gram features of instructions.
We configure the signature size to 128 and the similarity threshold to 0.7.
The final synthesized dataset comprises 100,000 instruction-response pairs.

To evaluate the effectiveness of WebR in generating high-quality IT datasets, we use WebR to construct datasets with two LLMs: the open-source \texttt{Llama3-70B-Instruct}~\cite{llama3modelcard} (temperature=0.6, top-p=0.9) and the proprietary \texttt{GPT-4o-mini}~\cite{2023gpt4} (temperature=0.7, top-p=1.0).
The resulting datasets, \textbf{WebR-Basic} (from Llama3) and \textbf{WebR-Pro} (from GPT-4o-mini), differ in their generative capabilities and quality.
A comparative analysis of the average token lengths is presented in Appendix \ref{sec:data analysis}, while a detailed cost analysis of WebR is provided in Appendix \ref{sec:cost}.
Notably, the overall expenditure for calling GPT-4o-mini API is \textbf{\$38.57}.

\begin{table*}[t]
  \footnotesize
  \centering
  \begin{tabularx}{\textwidth}{l c >{\centering\arraybackslash}X c | >{\centering\arraybackslash}X >{\centering\arraybackslash}X >{\centering\arraybackslash}X c c >{\columncolor{gray!15}} c}
    \toprule
     & & \textbf{Human} & \textbf{Response}  & \textbf{Alpaca} & \textbf{Arena} & \textbf{MT} & \multicolumn{2}{c}{\textbf{IFEval}} & \\
    \multirow{-2}{*}{\textbf{IT Data}} & \multirow{-2}{*}{\textbf{\#Data}} & \textbf{Effort} & \textbf{Generator} &\textbf{Eval 2} &\textbf{Hard} &\textbf{Bench} &\textbf{Pr. (S}) &\textbf{Ins. (S)} & \multirow{-2}{*}{\textbf{Avg.}}\\
    \midrule
        None (w/o fine-tuning) & - & - & - & 0.18 & 0.31	&1.78	&16.26	&18.01	&7.31 \\
        \midrule
        ShareGPT &  112k &High & ChatGPT  & 9.89 & 6.49 & 6.34 & 38.52 & 42.26 & 22.70                      \\
        WildChat &  652k &High & GPT-3.5 \& 4     & 14.62 & 8.73 & 6.60 & 39.53  & 45.66  &23.03   \\
        Tulu V2 Mix & 326k &Mid & Mix  & 9.91 & 5.41 & 5.76  & 37.69 & 41.05 &19.96       \\
        OpenHermes 2.5 & 1M &Mid & Mix   & 12.89 & 8.20 & 6.51  & 38.82 & 43.52 &21.99       \\
        Alpaca &  52k & Low & Davinci-003          & 4.21 & 1.24 & 3.75 & 20.21  & 23.56 &10.59                      \\
        Evol Instruct & 143k &Low & ChatGPT  & 7.19 & 5.58 & 5.77 & 39.00  & 44.25 &20.36                       \\
        WebInstruct & 100k &No & Llama3-70B  & 3.43 & 1.69 & 5.35  & 18.99 & 20.56 & 10.00 \\ 
        Backtranslation & 100k &No & Llama3-70B  & 5.24 & 2.81 & 3.74  & 26.85 & 29.61 & 13.65 \\
        DoG-Instruct & 100k & No & Llama3-70B  & 11.75 & 8.07 & 5.92 & 36.60  & 41.87 & 20.84 \\
        Magpie & 100k &No & Llama3-70B  & 23.62 & 13.98 & 6.26 & 33.83 & 43.07 & 24.15\\
        WebR-Basic & 100k &No & Llama3-70B  & \textbf{25.33} & \textbf{16.50} & \textbf{6.95} & \textbf{41.40} & \textbf{50.69} &\textbf{28.17}\\ \midrule
         IT Mix & 100k &Mid & GPT-4o-mini  & 30.39 & 28.03 & 7.36 & 43.30 & 47.38 &31.29\\
         Magpie & 100k &No & GPT-4o-mini  & 32.61 & 27.97 & 7.26 & 36.81 & 45.07 & 29.95 \\
        WebR-Pro & 100k &No & GPT-4o-mini  & \textbf{34.36} & \textbf{31.10} & \textbf{7.57} & \textbf{43.79} & \textbf{51.76} &\textbf{33.71}\\ \midrule
        (IT + WebR-Pro) Mix & 100k & Mid & GPT-4o-mini & 35.00 & 34.23 & 7.50 & 48.06 & 53.23 &35.60\\
        (IT + WebR-Pro) Merge & 200k & Mid & GPT-4o-mini & \textbf{35.40} & \textbf{35.12} & \textbf{7.59} & \textbf{49.72} & \textbf{53.97} &\textbf{36.36}\\
    \bottomrule
  \end{tabularx}
  \caption{
    Instruction-following performance comparison of various instruction-tuning (IT) data, based on Llama3-8B.
  }
  \label{tab:main_table}
\end{table*}

\section{Experimental Setup}

\subsection{Baselines}
We compare the family of IT datasets generated by WebR with ten state-of-the-art (SOTA) open-source IT datasets, categorized as follows:
(1) Human-crafted data: \textbf{ShareGPT}~\cite{vicuna2023} and \textbf{WildChat}~\cite{zhao2024wildchat} are exemplary human-written datasets containing 112K and 652K high-quality multi-round conversations between humans and GPT, respectively.
(2) Semi-automated synthetic data: \textbf{Alpaca}~\cite{alpaca} and \textbf{Evol-Instruct}~\cite{xu2024wizardlm} represent widely-used synthetic datasets generated with semi-automated techniques.
(3) Mixed data: \textbf{Tulu V2 Mix}~\cite{ivison2023camelschangingclimateenhancing} and \textbf{OpenHermes 2.5}~\cite{OpenHermes} are crowd-sourced datasets that aggregate diverse open-source IT datasets, featuring 326K and 1M conversations, respectively.
(4) Fully automated synthetic data: 
\textbf{Magpie}~\cite{xu2024magpie} synthesizes IT data by prompting Llama3-70B-Instruct with its chat template, from which we sample 100k examples.
To ensure a fair and controlled comparison, we reproduce several representative web-based IT synthesis methods—namely \textbf{WebInstruct} \cite{yue2024mammoth2}, \textbf{Backtranslation} \cite{li2024selfalignment}, and \textbf{DoG-Instruct}~\cite{chen-etal-2024-dog}—\textit{using the same source web data} as our proposed WebR. All methods are implemented based on the LLaMA3-70B-Instruct model, thereby aligning model capacity and input sources across approaches.

\subsection{Models and Training Settings}
For instruction tuning (IT), we train Llama3-8B-base~\cite{llama3modelcard} and Qwen2.5-1.5/3/7/14B-base~\cite{qwen2.5} on various IT datasets.
For each IT dataset, we fine-tune models with five different random seeds and report the average performance.
We adhere to the official instruction templates provided by each model.
To ensure a fair comparison, we use consistent training hyperparameters across different baseline datasets.
The comprehensive implementation details are listed in Appendix \ref{sec:appendix_implementation}.

\subsection{Evaluation Benchmarks and Metrics}
We evaluate the performance of the fine-tuned models using four widely adopted instruction-following benchmarks: AlpacaEval 2~\cite{li2023alpacaeval}, Arena-Hard~\cite{li2024live}, MT-Bench~\cite{zheng2023judging}, and IFEval~\cite{zhou2023ifeval}.
For AlpacaEval 2, we report the length-controlled win rate (LC), which ensures robustness against verbosity.
For Arena-Hard, we report the win rate (WR) against the baseline model.
For MT-Bench, we provide the average score, using GPT-4-turbo as the evaluation judge.
For IFEval, we report two metrics: prompt-level strict accuracy (\textit{Pr. (S)}) and instruction-level strict accuracy (\textit{Ins. (S)}).
More evaluation details are listed in Appendix \ref{sec:evaluation_details}.

\section{Experimental Results}

\subsection{Main Results}

\begin{table*}[t]
  \footnotesize
  \centering
  \begin{tabularx}{\textwidth}{l|>{\centering\arraybackslash}X|>{\centering\arraybackslash}X c | >{\centering\arraybackslash}X >{\centering\arraybackslash}X | c | >{\columncolor{gray!15}}>{\centering\arraybackslash}X}
    \toprule
     \textbf{IT Data} & \textbf{MMLU}  & \textbf{ARC} & \textbf{WinoGrande} & \textbf{MATH} & \textbf{GSM8K} & \textbf{HumanEval} & \textbf{Avg.} \\
    \midrule
    None (w/o fine-tuning) & 60.56	&73.52	&52.14	&19.62	&56.16	&39.08	&50.18 \\
    \midrule
    WildChat & 58.46 & 72.62 & 49.43 & 19.34 & 60.25 & 42.55 & 50.44 \\
    OpenHermes 2.5 & 60.08 & 75.65 & 51.22 & 24.18 & 64.70 & 44.43 & 53.38 \\
    Magpie & 58.58 & 71.53 & 51.93 & 16.12 & 57.39 & 40.85 & 49.40 \\
    WebR-Basic & 60.85 & 76.27 & 52.91 & 20.28 & 55.57 & 40.10 & 51.00 \\
    IT Mix & 57.44 & 73.56 & 50.36 & 22.00 & 61.87 & 45.12 & 51.73 \\
    WebR-Pro & \textbf{61.15} & 74.92 & \textbf{53.20} & 24.94 & 60.69 & 48.73 & 53.94 \\
    (IT + WebR-Pro) Mix & 60.69 & \textbf{77.63} & 50.67 & 26.34 & 64.90 & \textbf{50.61} & 55.14 \\
    (IT + WebR-Pro) Merge & 61.02 & 76.27 & 52.72 & \textbf{28.36} & \textbf{66.41} & \textbf{50.61} & \textbf{55.90} \\
    \bottomrule
  \end{tabularx}
  \caption{
    Performance comparison of downstream tasks (Knowledge, Reasoning, Math, Code) based on Llama3-8B.
  }
  \label{tab:capabilities}
\end{table*}

\begin{table*}[t]
  \footnotesize
  \centering
  \begin{tabularx}{\textwidth}{l|>{\centering\arraybackslash}X >{\centering\arraybackslash}X >{\centering\arraybackslash}X >{\columncolor{gray!15}}>{\centering\arraybackslash}X | >{\centering\arraybackslash}X >{\centering\arraybackslash}X >{\centering\arraybackslash}X c >{\columncolor{gray!15}}>{\centering\arraybackslash}X}
    \toprule
     & \textbf{Alpaca} & \textbf{MT} & \textbf{IFEval} & & & & & & \\
     \multirow{-2}{*}{\textbf{Setting}} &\textbf{Eval 2} &\textbf{Bench} &\textbf{Pr. (S}) & \multirow{-2}{*}{\textbf{Avg.}} & \multirow{-2}{*}{\textbf{MMLU}} & \multirow{-2}{*}{\textbf{ARC}} & \multirow{-2}{*}{\textbf{MATH}} & \multirow{-2}{*}{\textbf{HumanEval}} & \multirow{-2}{*}{\textbf{Avg.}} \\
    \midrule
    WebR-Pro & 34.17 & 7.50 & 43.55 & \textbf{28.41} & 61.15 & 74.92 & 24.94 & 48.73 & \textbf{52.43} \\ \midrule
    -w/o Persona & 33.30 & 6.93 & 44.69 & 28.31 & 60.98 & 74.58 & 24.03 & 48.50 & 52.02  \\
    -w/o Part & 33.89 & 7.53 & 42.60 &28.01 & 61.05 & 72.53 & 22.73 & 48.41 & 51.18   \\
    -w/o Refinement & 31.61 & 7.42 & 44.73 &27.92 & 59.83 & 74.92 & 24.36 & 48.61 & 51.93 \\
    -w/o MinHash & 32.43 & 7.29 & 43.02 &27.58& 60.69 & 74.92 & 24.82 & 47.15 & 51.90 \\ \midrule
    \multicolumn{9}{c}{\textit{Ratio of Web as Instruction to Web as Response (2 : 1 in WebR)}} \\ 
    1 : 0 & 29.15 & 7.10 & 39.56 &25.27 & 58.79 & 74.58 & 25.74 & 50.00 & 52.28 \\
    1 : 1 & 33.16 & 7.39 & 43.26 &27.94 & 60.60 & 73.22 & 25.18 & 48.78 & 51.95 \\
    1 : 2 & 32.99 & 7.33 & 42.85 &27.72 & 57.76 & 72.61 & 25.26 & 50.00 & 51.41 \\
    0 : 1 & 33.41 & 6.68 & 42.54 &27.54 & 52.68 & 72.90 & 23.30 & 46.95 & 48.96 \\
    \bottomrule
  \end{tabularx}
  \caption{
    Ablation study based on Llama3-8B.
  }
  \label{tab:ablation}
\end{table*}

\paragraph{WebR Outperforms Existing Baselines.}
Table \ref{tab:main_table} highlights the performance of Llama3-8B-base fine-tuned with datasets generated by WebR, compared to those fine-tuned with baseline datasets.
A general trend emerges: IT datasets requiring higher human effort tend to exhibit better performance than those with lower or no human effort.
Nevertheless, our WebR-Basic, which entirely eliminates human effort in dataset creation, significantly and consistently surpasses the SOTA Magpie dataset across all four benchmarks with a \textbf{16.65\%} average improvement.
To ensure a fair and more challenging comparison, we deduplicated and randomly sampled 100k instructions from baseline datasets of varying human effort levels (high, mid, and low) and generated responses using GPT-4o-mini, naming this synthesized strong baseline "\textbf{IT Mix}."
We also generate responses using GPT-4o-mini for Magpie and compare with our proposed method.
Even under the same response generator, WebR-Pro consistently outperforms IT Mix and Magpie by \textbf{7.73\%} and \textbf{12.55\%}, respectively. These results validate that datasets generated by WebR possess superior quality, enabling significantly enhanced instruction-following performance.

\paragraph{Compatibility of WebR.}
To explore the potential synergy between WebR and existing datasets, we merged IT Mix and WebR-Pro using two strategies: (1) random sampling of 50k data points from each dataset and (2) direct concatenation.
As shown in Table \ref{tab:main_table}, both merged datasets deliver further performance improvements over their individual components, establishing new SOTA results.
This can be attributed to the complementary strengths of the datasets: IT Mix offers broader data coverage, while WebR-Pro provides higher quality and more challenging instructions, as evidenced in Figure \ref{fig:quality_difficulty}.

\paragraph{Performance on Downstream Benchmarks.}
We evaluate the impact of various instruction-tuning datasets on downstream task performance across multiple domains\footnote{Evaluation settings are aligned with \url{https://opencompass.org.cn}.}:
(1) \textbf{Knowledge}: MMLU~\cite{DBLP:conf/iclr/HendrycksBBZMSS21};
(2) \textbf{Reasoning}: ARC~\cite{clark2018think} and WinoGrande~\cite{sakaguchi2019winogrande};
(3) \textbf{Math}: MATH~\cite{hendrycksmath2021} and GSM8K~\cite{cobbe2021gsm8k};
(4) \textbf{Code}: HumanEval~\cite{chen2021codex}.
As shown in Table~\ref{tab:capabilities}, models fine-tuned on the WebR datasets outperform those trained on other baselines, demonstrating their effectiveness in improving generalization across diverse downstream tasks, especially in challenging benchmarks like ARC and WinoGrande.
Furthermore, the combination of WebR-Pro and IT Mix further validates the complementary strengths of WebR data in aligning models with complex task requirements.

\subsection{Ablation Study}
Table \ref{tab:ablation} compares the LLM performance using different settings to construct WebR-Pro.

\begin{itemize}[noitemsep, topsep=0pt]
\item \textbf{w/o Persona}: removing the author's persona information during instruction generation leads to performance declines across almost all benchmarks.
\item \textbf{w/o Part}: creating instructions solely from the entire web content, rather than using specific parts, causes notable performance degradation, particularly on IFEval and reasoning-intensive tasks like ARC and MATH.
\item \textbf{w/o Refinement}: skipping the refinement step for \textit{Web as Response}—by directly adopting the rollout response as the final output—results in a substantial drop in instruction-following performance.
\item \textbf{w/o MinHash}: eliminating MinHash-based deduplication decreases performance across all benchmarks, highlighting the importance of maintaining dataset diversity.
\item \textbf{Ratio of \textit{Web as Instruction} to \textit{Web as Response}}: varying the ratio of \textit{Web as Instruction} to \textit{Web as Response} data synthesis reveals that \textbf{each component contributes uniquely to model capabilities}.
Specifically, \textit{Web as Instruction} enhances reasoning and understanding tasks (e.g., ARC and MATH), while \textit{Web as Response} primarily improves instruction-following and question-answering tasks (e.g., IFEval and AlpacaEval 2).
The optimal balance is achieved at a ratio of 2:1, which delivers the best overall performance.
\end{itemize}

\section{Analysis}

\subsection{Dataset Analysis of WebR}

\begin{table}[!t]
\centering
\footnotesize
\begin{tabular}{lccc}
\toprule
\textbf{IT Data} & \textbf{Human Effort} & \textbf{Avg. Score} & \textbf{Diversity} \\
\midrule
WildChat & high & 23.03 & \textbf{0.93} \\
OpenHermes & mid & 21.99 & 0.87 \\
Evol Instruct & low & 20.36 & 0.88 \\
WebInstruct & no & 9.79 & 0.84 \\
Magpie & no & 24.15 & 0.92 \\
WebR-Basic & no & 28.17 & 0.91 \\
WebR-Pro & no & \textbf{33.58} & \textbf{0.93} \\
\bottomrule
\end{tabular}
\caption{Comparison of embedding diversity.}
\label{tab:it_data_comparison}
\end{table}

\begin{figure}[!t]
\centering
\includegraphics[width=\linewidth]{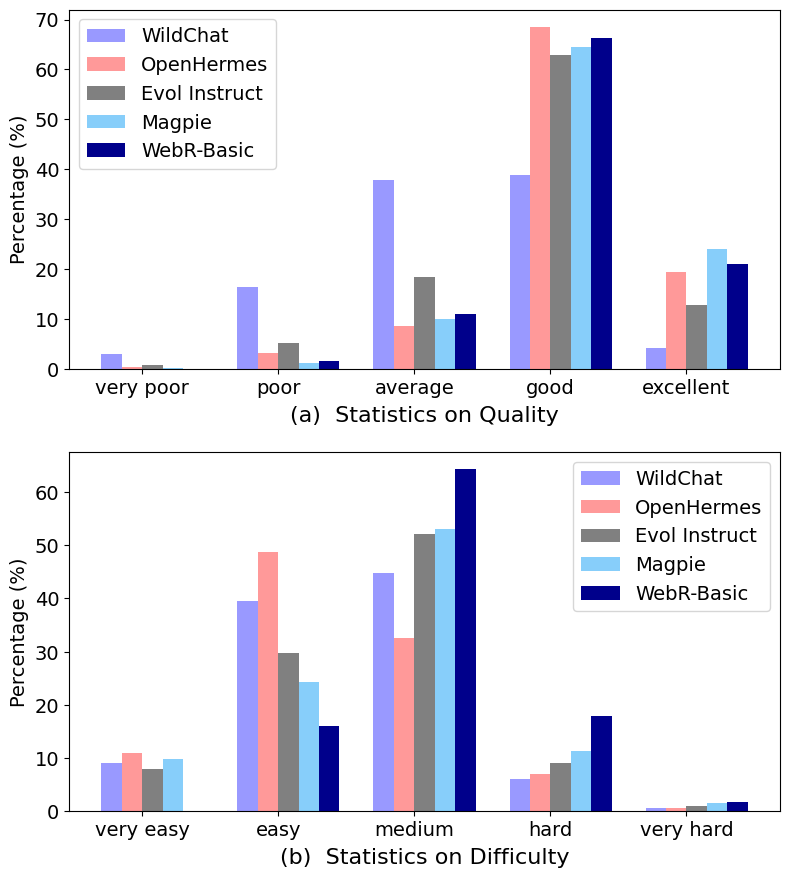}
\caption{
Statistics of instruction quality and difficulty. 
}
\label{fig:quality_difficulty}
\end{figure}

\paragraph{Diversity.}
We utilize a quantitative measure of diversity: (1) we randomly sample $N=10,000$ instructions from each dataset and encode them using the \texttt{all-mpnet-base-v2}\footnote{\url{https://huggingface.co/sentence-transformers/all-mpnet-base-v2}} embedding model; (2) we compute the average cosine similarity between all embedding pairs and define embedding diversity as $1- \frac{1}{C(N, 2)} \sum_{\forall i < j} \cos(\mathbf{e}_i, \mathbf{e}_j)$, where higher values indicate greater diversity.
Table \ref{tab:it_data_comparison} demonstrates that WebR-Pro achieves the highest diversity score (0.93), matching that of WildChat, which involves high human effort. Notably, WebR-Pro surpasses all other datasets—including those requiring human annotation like OpenHermes (0.87) and Evol Instruct (0.88)—indicating its strong capability to generate diverse instructions automatically. Furthermore, it outperforms previous automatic baselines such as WebInstruct (0.84) and Magpie (0.92), highlighting its effectiveness in promoting diversity without human intervention.



\paragraph{Quality and Difficulty.}
Following Magpie~\cite{xu2024magpie}, we use the Qwen2.5-72B-Instruct model to evaluate the quality and difficulty of each instruction, categorizing them into five levels. 
As depicted in Figure \ref{fig:quality_difficulty}, synthetic data generally demonstrates higher quality and greater difficulty compared to human-crafted instructions.
In particular, WebR-Basic exhibits superior distributions in both quality and difficulty metrics, surpassing existing baselines in these aspects.

\begin{figure}[!t]
\centering
\includegraphics[width=\linewidth]{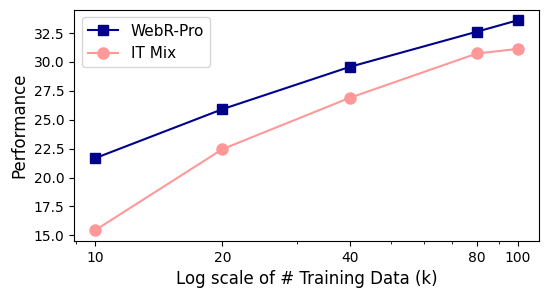}
\caption{
The impact of training data scale on the average instruction-following performance.
}
\label{fig:data_scale}
\end{figure}

\begin{table*}
  \footnotesize
  \centering
  \begin{tabularx}{\textwidth}{l l >{\centering\arraybackslash}X >{\centering\arraybackslash}X >{\centering\arraybackslash}X >{\centering\arraybackslash}X >{\centering\arraybackslash}X}
    \toprule
     \textbf{Base LLM} & \textbf{IT Data} & \textbf{AlpacaEval 2} & \textbf{Arena-Hard} & \textbf{MT-Bench} & \textbf{IFEval/Pr. (S)} & \textbf{IFEval/Ins. (S)} \\
    \midrule
    \multirow{2}{*}{Qwen2.5-1.5B}
         & IT Mix &10.98  &\textbf{15.10}  & \textbf{6.03} &\textbf{29.57}  &\textbf{33.27}  \\
        & WebR-Pro  &\textbf{11.00} (\textcolor{customgreen}{+0.02})  & 14.03 (\textcolor{customred}{-1.07})  & 5.92 (\textcolor{customred}{-0.11}) & \textbf{29.57} (\textcolor{customgreen}{+0.00})  & 32.16 (\textcolor{customred}{-1.11}) \\
    \midrule
    \multirow{2}{*}{Qwen2.5-3B}
         & IT Mix &\textbf{22.36}  &26.54  & 6.95  &\textbf{43.07}  &\textbf{44.73}  \\
        & WebR-Pro  & 22.29 (\textcolor{customred}{-0.07})  &\textbf{28.13} (\textcolor{customgreen}{+1.59})  & \textbf{7.03} (\textcolor{customgreen}{+0.08})  &42.38 (\textcolor{customred}{-0.69})  &44.71 (\textcolor{customred}{-0.02}) \\
    \midrule
    \multirow{2}{*}{Qwen2.5-7B}
         & IT Mix &32.59  &45.10  &7.45  &49.35  &52.68  \\
        & WebR-Pro  &\textbf{34.90} (\textcolor{customgreen}{+2.31})  &\textbf{45.66} (\textcolor{customgreen}{+0.56})  &\textbf{7.62} (\textcolor{customgreen}{+0.17})  &\textbf{50.55} (\textcolor{customgreen}{+1.20})  &\textbf{53.35} (\textcolor{customgreen}{+0.67}) \\
    \midrule
    \multirow{2}{*}{Qwen2.5-14B}
         & IT Mix &42.07  &59.00  &8.10  &58.04  &60.63  \\
        & WebR-Pro  &\textbf{46.19} (\textcolor{customgreen}{+4.12})  &\textbf{62.13} (\textcolor{customgreen}{+2.13})  &\textbf{8.39} (\textcolor{customgreen}{+0.29})  &\textbf{60.23} (\textcolor{customgreen}{+2.19})  &\textbf{64.88} (\textcolor{customgreen}{+4.25}) \\
    \bottomrule
  \end{tabularx}
  \caption{
    Performance comparison across varied scales of base LLMs.
  }
  \label{tab:llm_size}
\end{table*}

\begin{table*}
  \footnotesize
  \centering
  \begin{tabularx}{\textwidth}{lc >{\centering\arraybackslash}X c >{\centering\arraybackslash}X >{\centering\arraybackslash}X >{\columncolor{gray!15}}>{\centering\arraybackslash}X}
    \toprule
    \textbf{Data Proportion} & \textbf{AlpacaEval 2}  & \textbf{MATH} & \textbf{HumanEval} & \textbf{MedQA} & \textbf{FinBen} & \textbf{Avg.} \\
    \midrule
    IT Mix & 30.19 & 22.00 & 45.12 & 38.88 & 29.20 & 33.08 \\ \midrule
    WebR-Pro (4.7 gen : 1 math : 1 code) & 34.17 & 24.94 & 48.73 & 47.31 & 29.56 & 36.94 \\
    - 1 gen & 34.40 & 22.52 & 44.78 & 44.94 & 28.97 & 35.12 \\
    - 1 gen : 1 math & 34.25 & \textbf{\textcolor{customgreen}{28.09}} & 48.23 & 46.59 & 29.77 & 37.39 \\
    - 1 gen : 1 math : 1 code & 34.59 & 27.10 & \textbf{\textcolor{customgreen}{51.39}} & 46.83 & 29.34 & \textbf{37.85} \\
    - 1 gen : 1 math : 1 code : 1 med &32.75 & 26.22 & 49.68 & \textbf{\textcolor{customgreen}{49.98}} & 29.01 & 37.53 \\
    - 1 gen : 1 math : 1 code : 1 med : 1 fin &33.03 & 25.38 & 48.17 & 45.64 & \textbf{\textcolor{customgreen}{30.22}} & 36.49 \\
    \bottomrule
  \end{tabularx}
  \caption{
    Domain adaptation based on Llama3-8B, with the domain improvements marked in \textcolor{customgreen}{\textbf{green}}.
  }
  \label{tab:domain}
\end{table*}

\subsection{Data Efficiency of WebR}
Figure \ref{fig:data_scale} illustrates the impact of training data scale on model performance.
The results clearly underscore the superior data efficiency of WebR-Pro compared to IT Mix:
(1) With only 10k training samples, WebR-Pro achieves a striking \textbf{40.26\%} performance improvement over IT Mix, highlighting its exceptional capability to elicit latent potential from LLMs even with limited data.
(2) WebR-Pro exhibits a more consistent and pronounced \textbf{linear} performance increase with respect to the logarithmic growth in training data, consistently outperforming IT Mix across all data scales.
These results strongly validate the efficacy of WebR in efficiently leveraging training data to unlock and enhance the capabilities of LLMs.

\subsection{Scalability of WebR}
Table \ref{tab:llm_size} highlights the impact of base LLM scale on the performance of our proposed WebR method.
While WebR-Pro slightly underperforms IT Mix at the 1.5B model scale, its advantages become increasingly pronounced as the model size grows.
For instance, WebR-Pro achieves an average performance improvement of \textbf{2.86\%} over IT Mix with Qwen2.5-7B and an even more substantial improvement of \textbf{5.55\%} with Qwen2.5-14B.
These results suggest that the advanced synthesis paradigm of WebR better aligns with larger models' capacity to capture complex patterns and utilize reasoning-intensive data.
In contrast, smaller models with limited capacity may struggle to fully exploit WebR’s potential.

\subsection{Domain Adaptability of WebR}
We explore the potential of our proposed WebR framework for domain adaptation by simply adjusting the proportion of source web documents.
Starting with general-domain content, we progressively add domain-specific materials from math, code, medicine, and finance, assessing performance across relevant benchmarks.
For the medical and financial domains, we utilize raw web documents from IndustryCorpus2~\cite{beijing_academy_of_artificial_intelligence}, and evaluate using MedQA~\cite{jin2021medqa} and FinBen~\cite{xie2024finben} benchmarks.
As shown in Table~\ref{tab:domain}, WebR demonstrates strong adaptability across domains. Compared to the IT Mix baseline, incorporating domain-specific data consistently improves performance, with math and code data yielding significant gains in MATH (28.09) and HumanEval (51.39), and medical and financial domains showing strong results on MedQA (49.98) and FinBen (30.22). These results highlight WebR’s ability to \textbf{incorporate specialized knowledge while maintaining competitive general-domain performance}.
Furthermore, the process of collecting domain-specific web documents is straightforward, underscoring WebR’s practicality.

\section{Conclusion}
In this paper, we present \textbf{Web Reconstruction} (WebR), a fully automated framework for synthesizing high-quality instruction-tuning (IT) datasets.
Harnessing the richness of raw web content, we conceptualize \textit{web reconstruction} as an instruction-tuning data synthesis task via a novel dual-perspective paradigm—\textit{Web as Instruction} and \textit{Web as Response}—where each web document is designated as either an instruction or a response to trigger the reconstruction process.
Extensive experiments show that WebR-generated datasets consistently outperform state-of-the-art baselines across four instruction-following benchmarks and six diverse downstream tasks.
Furthermore, WebR exhibits exceptional compatibility, data efficiency, and scalability with existing datasets, underscoring its potential as a versatile tool for advancing instruction-driven LLM training.

\section*{Limitations}
While WebR can already obtain satisfactory performance, there are several areas for improvement and future exploration.
Firstly, the current implementation of WebR focuses on single-turn data synthesis. Expanding this framework to support multi-turn conversations could further enhance its applicability to complex, interactive tasks.
Secondly, due to constraints in time and computational resources, the size of the constructed WebR-Basic and WebR-Pro datasets is currently limited to 100k samples.
However, given the vast availability of web documents—numbering in the trillions—the WebR framework has significant potential for scaling to create large-scale IT datasets, which could further boost performance.
Thirdly, WebR does not incorporate advanced data selection techniques, such as Instruction Following Difficulty (IFD)~\cite{li-etal-2024-quantity}, as part of its post-processing pipeline.
Incorporating such techniques could refine data quality and further enhance the instruction-following capabilities of LLMs \cite{DBLP:conf/acl/Jiang0ZZLMS00W24}.
Finally, advanced post-training techniques such as reinforcement learning from human feedback (RLHF)~\cite{ouyang2022training, huang-etal-2025-evolution} and direct preference optimization (DPO)~\citep{rafailov2024dpo, jiang2025bridging} can be applied to the synthesized WebR data to further enhance alignment performance, which we leave for future work.

\section*{Ethics Statement}
This study adheres strictly to the ethical principles established by the research community.
The utilized IT datasets are reported to be safe and free of content that may contain discrimination, personally
identifiable information, or any other undesirable
behaviors.
We have meticulously designed and curated our instructions for LLMs to ensure that all tasks are restricted to web reconstruction.
This approach minimizes the risk of generating content that could raise ethical concerns.

\section*{Acknowledgments}
Wei Wang was supported by the Guangdong Provincial Key Laboratory of Integrated Communication, Sensing, and Computation for Ubiquitous Internet of Things (Grant No. 2023B1212010007), the Guangzhou Municipal Science and Technology Project (Grant Nos. 2023A03J0003, 2023A03J0013, and 2024A03J0621), and the Institute of Education Innovation and Practice Project (Grant Nos. G01RF000012 and G01RF000017).

\bibliography{custom}

\clearpage

\appendix

\section{Implementation Details}
\label{sec:appendix_implementation}
Our implementation is based on the alignment-handbook repo\footnote{\url{https://github.com/huggingface/alignment-handbook}}.
The training procedure was executed on 4 NVIDIA A800 GPUs, each equipped with 80GB of memory.
The duration required to train a single instance of the model, specifically the Llama3-8B-base, was approximately 9 hours.
The specific hyperparameters used during training are detailed in Table \ref{tab:hyperparameters}.
Notably, all models were trained using the same set of hyperparameters, except for the maximum sequence length, which was set to 2048 for the 14B LLMs to mitigate computational bottlenecks.

\begin{table}[!h]
\centering
{\begin{tabular}{l|c}
\toprule
\textbf{Hyperparameter} & \textbf{Value} \\ \midrule
Batch size              & 128                            \\
Learning rate           & 2e-5                         \\
Epoches                 & 4                              \\
Max length              & 4096 (2048 for 14B LLMs)                          \\
Optimizer               & AdamW                     \\
Scheduler               & cosine             \\
Weight decay            & 0                       \\
Warmup ratio            & 0.1                  \\ \bottomrule
\end{tabular}}
\caption{\label{tab:hyperparameters}
Training hyperparameters for Llama3-8B-base and Qwen2.5-1.5/3/7/14B-base.}
\end{table}

\begin{table*}[t!]
\centering
\footnotesize
\begin{tabular}{lccccc}
\toprule
\textbf{Benchmark} & \textbf{\# Exs.} & \textbf{Baseline Model} & \textbf{Judge Model} & \textbf{Scoring Type} & \textbf{Metric} \\
\midrule
AlpacaEval 2 & 805 & GPT-4 Turbo & GPT-4 Turbo & Pairwise comparison & Length-controlled win rate \\
Arena-Hard   & 500 & GPT-4-0314 & GPT-4 Turbo & Pairwise comparison & Win rate \\
MT-Bench     & 80  & -          & GPT-4/GPT-4 Turbo & Single-answer grading & Rating of 1-10 \\
IFEval & 541 & - & - & Rule-based verification & Accuracy \\
\bottomrule
\end{tabular}
\caption{Evaluation details for AlpacaEval 2~\cite{li2023alpacaeval}, Arena-Hard~\cite{li2024live}, MT-Bench~\cite{zheng2023judging}, and IFEval~\cite{zhou2023ifeval}. The baseline model refers to the model compared against.}
\label{tab:evaluation_details}
\end{table*}

\begin{table*}[t!]
\centering
\footnotesize
\begin{tabularx}{\textwidth}{l >{\centering\arraybackslash}X >{\centering\arraybackslash}X >{\centering\arraybackslash}X >{\centering\arraybackslash}X}
\toprule
&  & \textbf{Avg. Input} & \textbf{Avg. Output} &  \\ 
& \multirow{-2}{*}{\textbf{\# of Samples}} & \textbf{Token Length} & \textbf{Token Length} & \multirow{-2}{*}{\textbf{Cost (\$)}} \\
\midrule
Generate author's persona & 100,000 & 523 & 32 & 4.88 \\
Web as Instruction (instruction) & 66,667 & 711 & 123 & 6.02 \\
Web as Instruction  (rollout response) & 66,667 & 611 & 392 & 10.90 \\
Web as Response (instruction) & 33,333 & 645 & 91 & 2.52 \\
Web as Response (rollout response) & 33,333 & 91 & 522 & 5.45 \\
Web as Response (refined response) & 33,333 & 1,155 & 591 & 8.80 \\ \midrule
Total & - & - & - & 38.57 \\
\bottomrule
\end{tabularx}
\caption{Estimated budget for data synthesis using the \texttt{GPT-4o-mini} API.}
\label{tab:api_cost}
\end{table*}


\section{Evaluation Details}
\label{sec:evaluation_details}
Table \ref{tab:evaluation_details} lists the evaluation details for AlpacaEval 2~\cite{li2023alpacaeval}, Arena-Hard~\cite{li2024live}, MT-Bench~\cite{zheng2023judging}, and IFEval~\cite{zhou2023ifeval}.
AlpacaEval 2 comprises 805 questions from 5 datasets, and MT-Bench spans 8 categories with a total of 80 questions.
Arena-Hard is an enhanced version of MT-Bench, featuring 500 well-defined technical problem-solving queries.
IFEval consists of 541 samples, each containing 1 to 3 verifiable constraints.
Evaluation metrics are reported in accordance with each benchmark's protocol.

\section{Dataset Analysis}
\label{sec:data analysis}
\begin{figure}[!t]
\centering
\includegraphics[width=\linewidth]{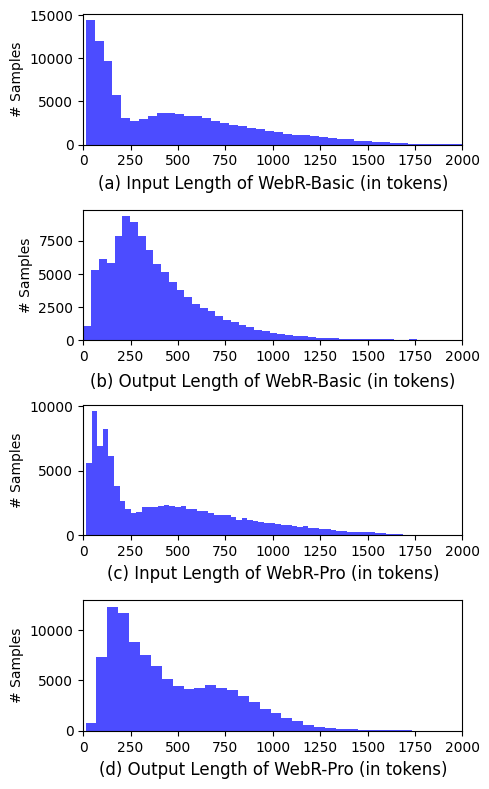}
\caption{
Lengths of instructions and responses in WebR-Basic and WebR-Pro. 
}
\label{fig:token_len}
\end{figure}

Statistics including token lengths of instructions and responses are illustrated in Figure \ref{fig:token_len}.
Tokens are counted using the \texttt{tiktoken} library\footnote{\url{https://github.com/openai/tiktoken}}.
For WebR-Basic, the average token lengths of instructions and responses are 441.41 and 381.28, respectively.
For WebR, the average token lengths of instructions and responses are 439.88 and 457.34, respectively.

\section{Cost Analysis}
\label{sec:cost}

Here we analyze the cost-effectiveness of our proposed Web Reconstruction framework.
For context, we estimated the budget for data synthesis using the \texttt{GPT-4o-mini} API, based on the Batch API's pricing of \$0.075 per 1M input tokens and \$0.3 per 1M output tokens.
Table \ref{tab:api_cost} lists the breakdown of the estimated costs for each step, which demonstrates that the overall expenditure (\textbf{\$38.57}) is both reasonable and manageable.

Additionally, our main experiment in Table \ref{tab:main_table} demonstrates that the open-source \texttt{Llama3-70B-Instruct} model can achieve satisfactory performance for our proposed Web Reconstruction, significantly outperforming previous IT datasets. 
Notably, \texttt{Llama3-70B-Instruct} can be deployed on only 2 NVIDIA-3090 GPUs, with the option to further reduce hardware requirements through low-bit quantization\footnote{\url{https://github.com/ollama/ollama}}.
This provides an economical alternative for our proposed WebR.
In conclusion, our framework demonstrates robustness in leveraging diverse LLMs for data synthesis, confirming its adaptability and effectiveness.

\section{Prompt Template}
\begin{figure*}[!ht]
\begin{center}
\includegraphics[width=\linewidth]{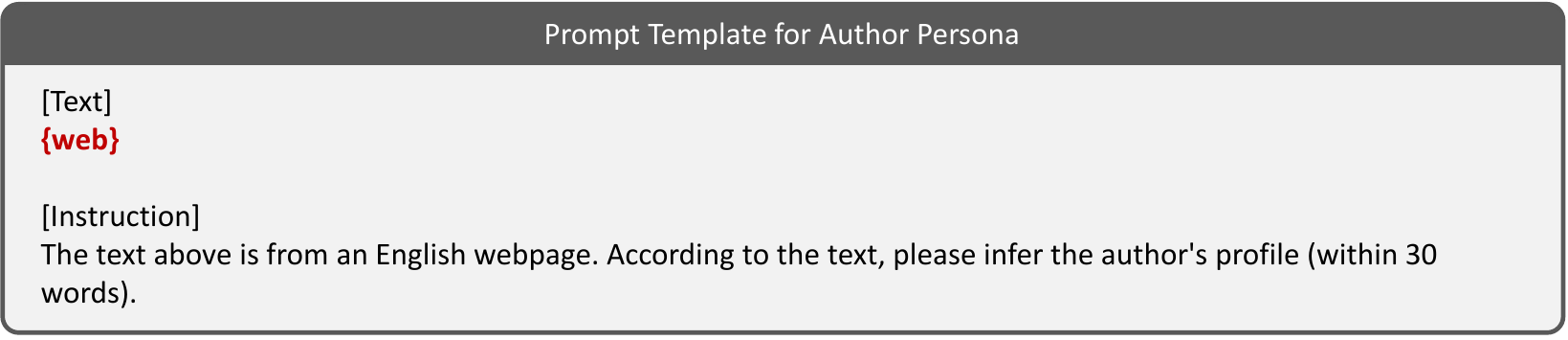}
\end{center}
\caption{Prompt template for generating author persona.}
\label{fig:persona_prompt}
\end{figure*}

\begin{figure*}[!ht]
\begin{center}
\includegraphics[width=\linewidth]{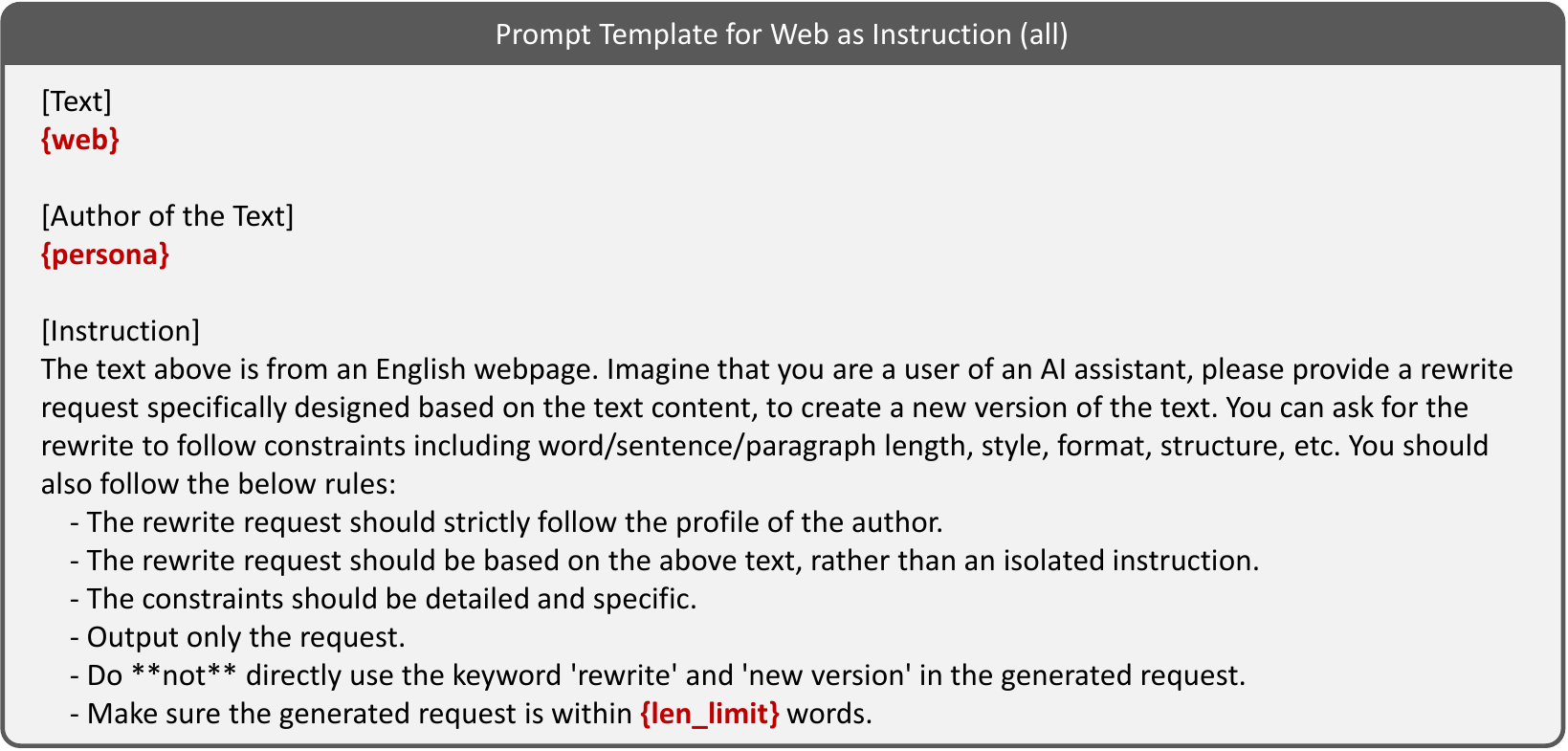}
\end{center}
\caption{Prompt template for \textit{Web as Instruction} (generating the rewrite request based on the whole web content).}
\label{fig:l2l_all_prompt}
\end{figure*}

\begin{figure*}[!ht]
\begin{center}
\includegraphics[width=\linewidth]{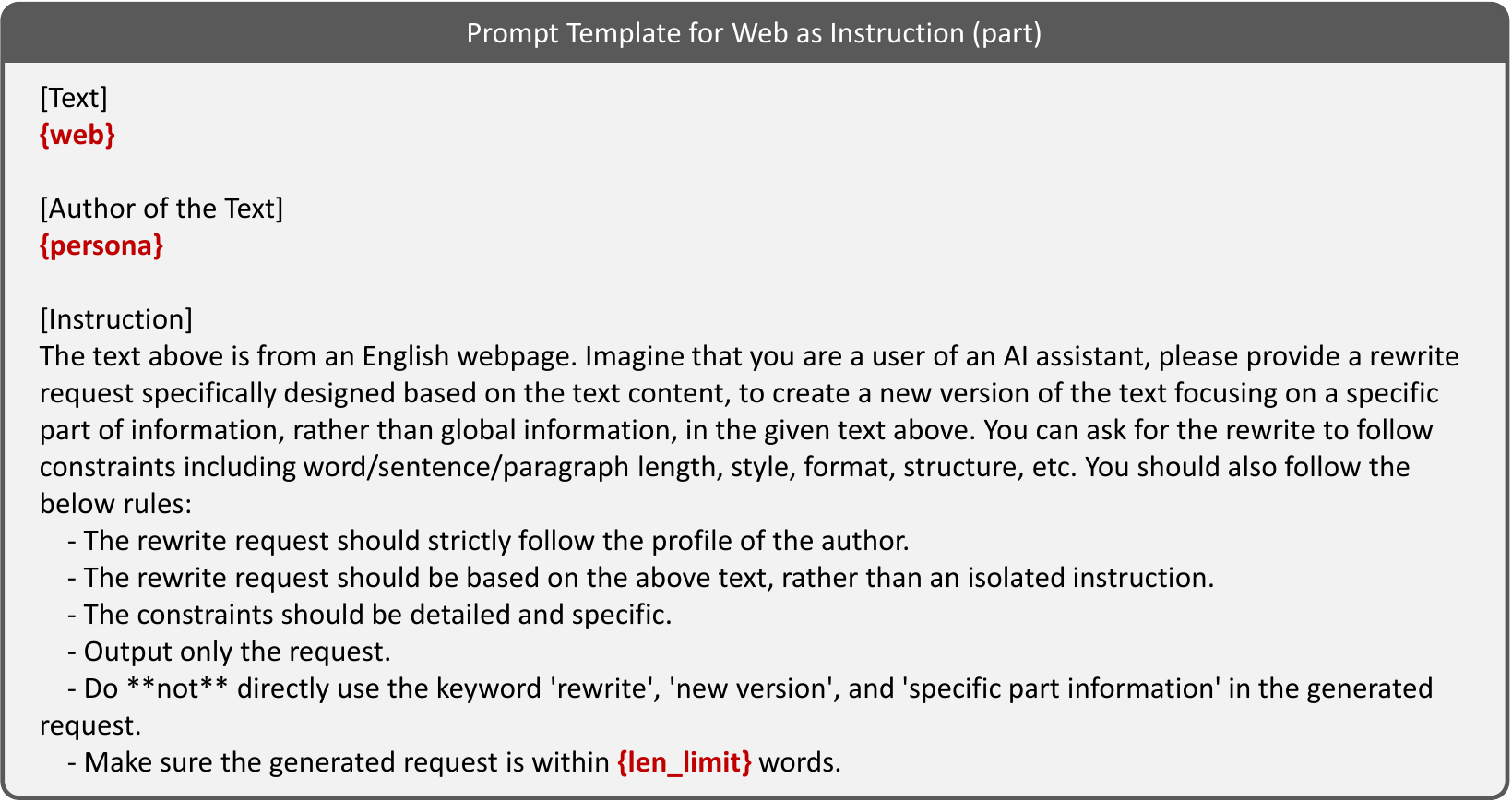}
\end{center}
\caption{Prompt template for \textit{Web as Instruction} (generating the rewrite request based on the specific part of the web content).}
\label{fig:l2l_part_prompt}
\end{figure*}

\begin{figure*}[!ht]
\begin{center}
\includegraphics[width=\linewidth]{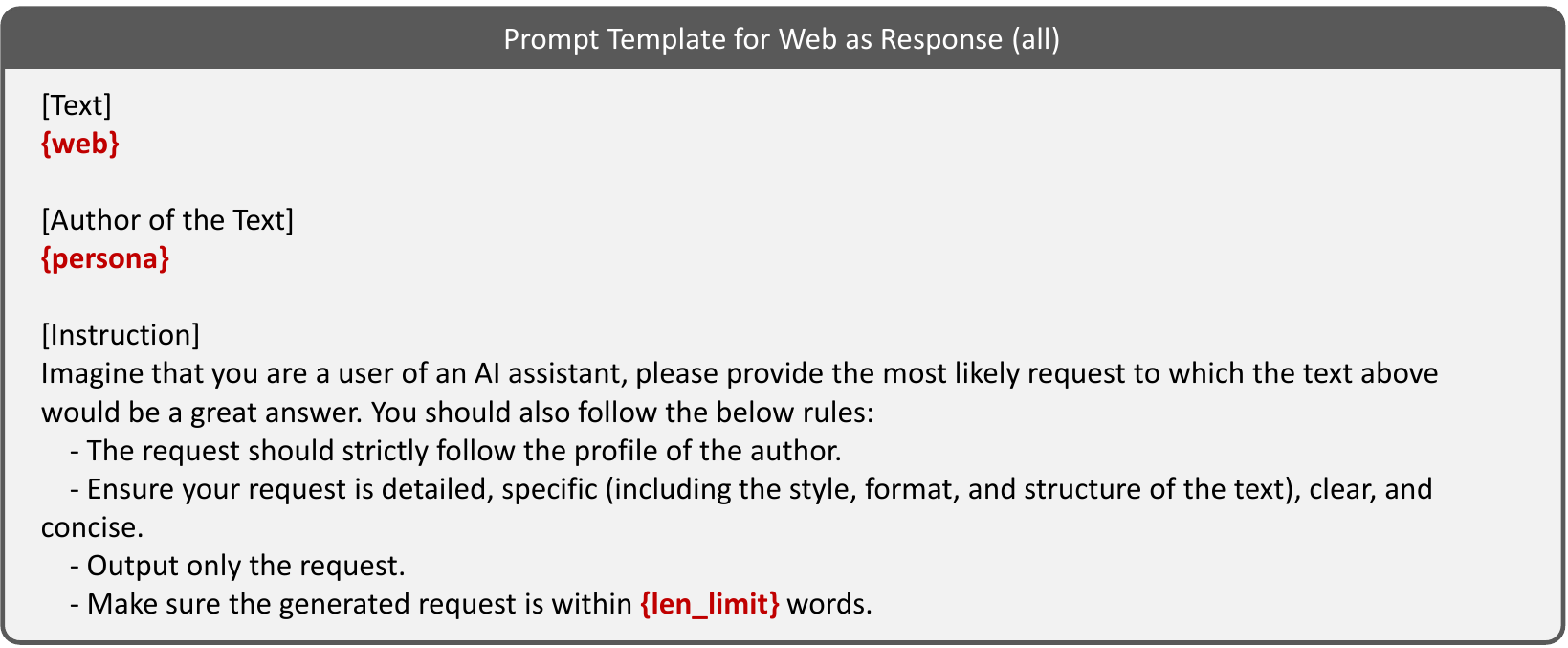}
\end{center}
\caption{Prompt template for \textit{Web as Response} (generating the latent instruction based on the whole web content).}
\label{fig:s2l_all_prompt}
\end{figure*}

\begin{figure*}[!ht]
\begin{center}
\includegraphics[width=\linewidth]{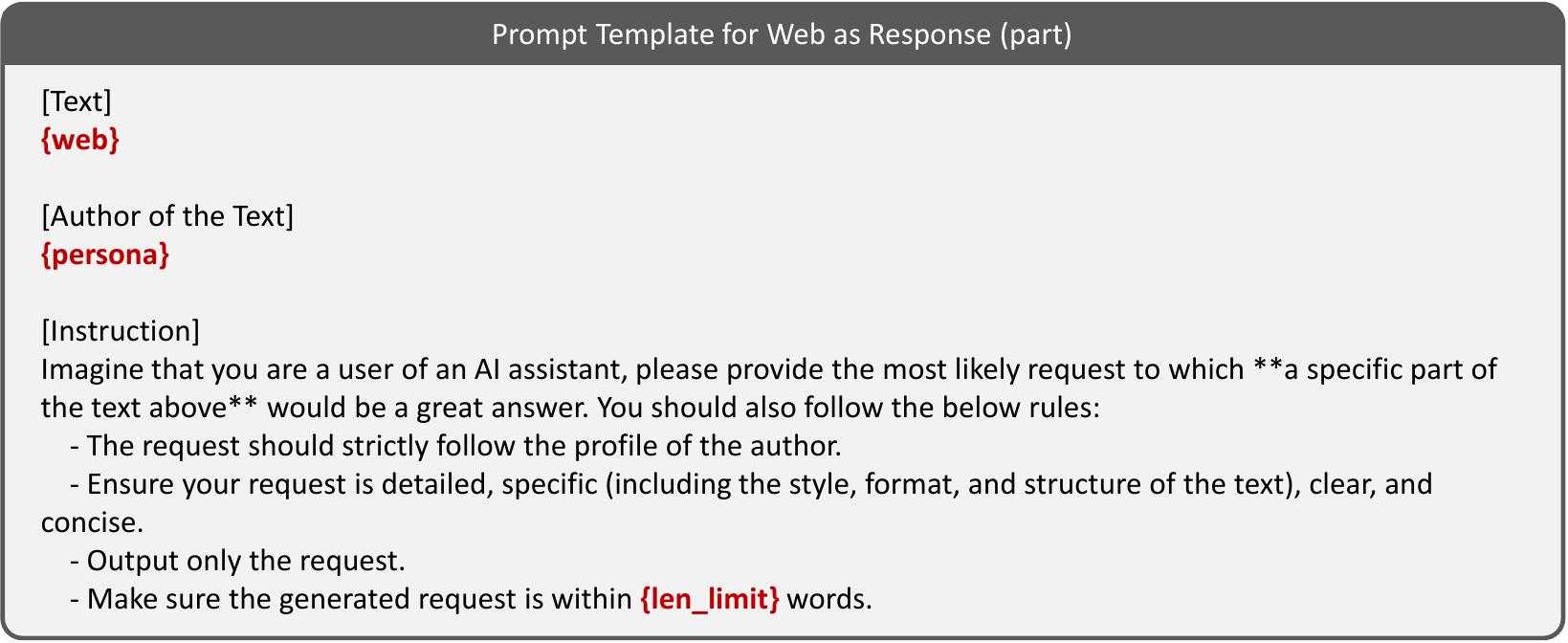}
\end{center}
\caption{Prompt template for \textit{Web as Response} (generating the latent instruction based on the specific part of the web content).}
\label{fig:s2l_part_prompt}
\end{figure*}

\begin{figure*}[!ht]
\begin{center}
\includegraphics[width=\linewidth]{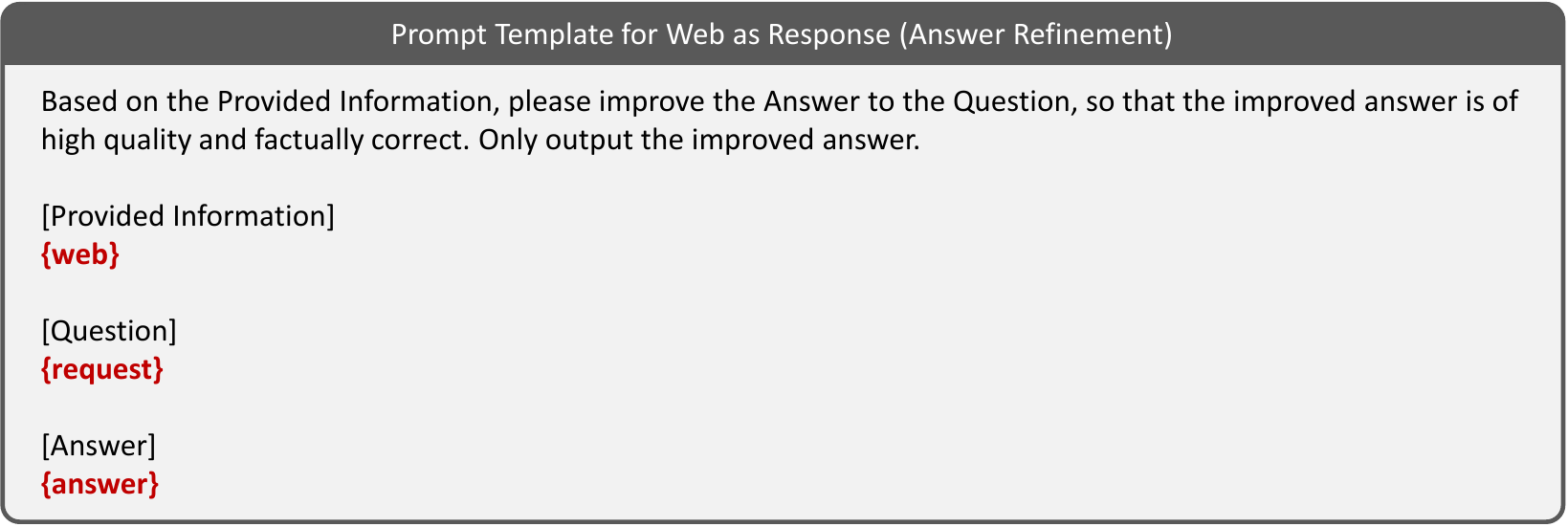}
\end{center}
\caption{Prompt template for \textit{Web as Response} (answer refinement).}
\label{fig:refine_prompt}
\end{figure*}

Figure \ref{fig:persona_prompt} shows the prompt template for generating the author persona according to the web content.
Figure \ref{fig:l2l_all_prompt} shows the prompt template for generating the rewrite request based on the whole web content.
Figure \ref{fig:l2l_part_prompt} shows the prompt template for generating the rewrite request based on a specific part of the web content.
Figure \ref{fig:s2l_all_prompt} shows the prompt template for generating the latent instruction corresponding to the whole web content.
Figure \ref{fig:s2l_part_prompt} shows the prompt template for generating the latent instruction corresponding to a specific part of the web content.
Figure \ref{fig:refine_prompt} shows the prompt template for generating a refined response based on the raw web and the instruction.

\end{document}